\documentclass[runningheads,a4paper]{llncs}
\usepackage{multirow}
\usepackage[english]{babel}

\usepackage{amsmath}
\usepackage{graphicx}
\usepackage{multirow}
\usepackage{float}
\usepackage{caption}
\usepackage{subcaption}
\usepackage{hyperref}
\usepackage{booktabs}

\graphicspath{{imgs/}}

\begin{document}

\title{Tumor Delineation For Brain Radiosurgery by a ConvNet and Non-Uniform Patch Generation}
\titlerunning{Tumor Delineation For Brain Radiosurgery}  
%
\author{
    Egor Krivov \inst{1, 2}  \and 
    Valery Kostjuchenko \inst{3}  \and 
    Alexandra Dalechina \inst{3}  \and 
    Boris Shirokikh \inst{4, 1, 2} \and
    Gleb Makarchuk \inst{4} \and
    Alexander Denisenko \inst{4}\and
    Andrey Golanov \inst{5} \and
    Mikhail Belyaev\inst{4, 1}
}

\authorrunning{E. Krivov et al}

\tocauthor{E. Krivov et al}

%

\institute{
    Kharkevich Institute for Information Transmission Problems, Moscow, Russia
    \and 
    Moscow Institute of Physics and Technology, Moscow, Russia
    \and 
    Moscow Gamma-Knife Center, Moscow, Russia
    \and
    Skolkovo Institute of Science and Technology, Moscow, Russia
    \and 
    Burdenko Neurosurgery Institute, Moscow, Russia
    \\
    \email{m.belyaev@skoltech.ru}
}

\maketitle              

\begin{abstract}
Deep learning methods are actively used for brain lesion segmentation. One of the most popular models is DeepMedic, which was developed for segmentation of relatively large lesions like glioma and ischemic stroke. In our work, we consider segmentation of brain tumors appropriate to stereotactic radiosurgery which limits typical lesion sizes. These differences in target volumes lead to a large number of false negatives (especially for small lesions) as well as to an increased number of false positives for DeepMedic. We propose a new patch-sampling procedure to increase network performance for small lesions.  We used a 6-year dataset from a stereotactic radiosurgery center. To evaluate our approach, we conducted experiments with the three most frequent brain tumors: metastasis, meningioma, schwannoma. In addition to cross-validation, we estimated quality on a hold-out test set which was collected several years later than the train one. The experimental results show solid improvements in both cases.
\end{abstract}

\section{Introduction}


During the last several years deep learning algorithms have gained a lot of attention from the academia since they showed previously unimaginable performance in various image analysis tasks. By now, deep learning methods are actively used in medical imaging as well \cite{litjens2017survey}. In particular,  deep convolutional networks dominate over traditional algorithms such as random forests in all recent MRI segmentation competitions (e.g., ischemic stroke  \cite{maier2017isles} or glioma \cite{brats1} segmentation).

However, we suppose that a gap exists between these results and MRI analysis in everyday clinical settings. The majority of open datasets for brain lesion segmentation are devoted to research-oriented questions such as ``is it possible to extract some biomarkers associated with the clinical outcome from lesion segmentation masks?'' \cite{brats2017}. Meanwhile, radiologists usually do not delineate lesions like glioma in their routine practice as it is a very time-consuming procedure and clinical protocols do not require it. We suppose that current deep learning-based algorithmic results lack verification in real-world clinical scenarios. Also, such verification can pose new specific requirements and therefore stimulate further algorithmic development.

In our work, we focus on adaptation of DeepMedic \cite{deepmedic_els}, a state-of-the-art deep learning convolutional network for brain lesion segmentation, for stereotactic radiosurgery. Delineation of pathological tissues is an obligatory part of radiosurgery planning and radiation oncologists have to detect and segment all tumors in MRI scans. So, radiosurgery is an interesting application area for deep learning methods \cite{zaharchuk2018deep}; recently two DeepMedic-based papers on brain metastasis segmentation were published \cite{liu2017deep,charron2018automatic}. We observe that the standard approach leads to a high number of false positives and propose a problem-oriented training procedure. 
To evaluate our approach, we use data on three most disseminated brain tumors (metastases, meningiomas, and acoustic schwannomas) \cite{gk_society_report} from a Gamma Knife radiosurgery center. We not only report  quality metrics for cross-validation, but also provide evaluation on a test set which was collected several years later than the training one to prove robustness of the developed models. To our knowledge, it is the first time, when modern deep learning algorithms were tested over such a long period of time in the field of MRI segmentation. 

\section{Related work}
\label{sec:related_works}

During the recent years, various deep learning architectures were developed. Unet, one of the most successful recent fully convolutional networks, was designed for 2D image segmentation. The core idea of the method is to add several additional connections between decoding and encoding paths to combine feature maps with various level of local and contextual information. For medical imaging, a straightforward 3D-convolutional generalization was proposed in \cite{cciccek20163d}. However, a large size of typical brain MR images place some restrictions on network receptive field. In such conditions, a more simple network called DeepMedic demonstrates solid performance in series of competitions, including glioma \cite{maier2017isles} and acute ischemic stroke segmentation \cite{brats1}.  The network is 11 layers deep and consists of two input paths: the first process a small patch of the image in the original resolution, the second one works with larger area in a coarser resolution and provides information on patch localization. The training is based on the following patch generation algorithm: the central voxel of each patch should belong to the target mask with predefined probability.

DeepMedic was recently used in two works on brain metastases segmentation. In \cite{liu2017deep} authors modified the original architecture by adding another input branch in original resolution and reported significant improvements over the original model. Their experiments with a metastases dataset resulted in Dice score equal to  $0.67$. The original DeepMedic architecture was used in \cite{charron2018automatic} where authors compared various combinations of T1c, T2 and Flair modalities. For T1c (in our paper we use only this modality) they reported Dice Score $0.77$, sensitivity $0.92$ and $10.5$ false positives per image. In both papers, the original patch generation strategy was used. 
A non-uniform patch generation process was proposed in \cite{ghafoorian2016non}. In fact, authors applied a predefined elastic deformation to each patch, whereas patch sampling (i.e. selection of the center voxel) was uniform.

\section{Data}
We focused on everyday practice of a radiosurgical center, that conducts operations with Leksell Gamma Knife. 
A typical Gamma Knife treatment consists of gathering patient data, frame fixation, performing an MRI scans, lesion delineation in MRI scans, treatment planning and, finally, the delivery of a dose of irradiation to a small intracranial volume through the intact skull. Delineation itself usually takes up to one hour. 

We found that three of the most popular diagnoses cover $77\%$ of all patient visits. This data is consistent with Leksell Gamma Knife society report \cite{gk_society_report}, according to which in 2016 metastases, meningioma and schwannoma accounted for $70\%$ cases treated in Gamma Knife centers. We focused only on these diagnoses since processing each new diagnosis makes our analysis more and more difficult, while clinical effect diminishes with the number of patients affected.

We use MRI T1c in all our experiments, image resolution is $(0.94, 0.94, 1)$ mm. We gathered two datasets from Gamma-Knife facility: historic and modern. Historic dataset consisted of patients examined between 2005 and 2011. Modern was gathered in 2017, we used it to ensure that developed methods could be used over a long period of time. Therefore, there was a 6 years gap between the last examination in the historic dataset and the first examination in modern dataset. Detailes are provided in Table~\ref{dataset_description}. Ground truth was provided by medical physicists who routinely perform tumor delineation procedure.

\vspace{-0.5em}

\begin{table}[]
\centering
\begin{tabular}{@{}lcccc@{}}
\toprule
                & Metastasis \hphantom & Meningioma \hphantom & Acoustic Schwannoma \\ \midrule
Historic \hphantom & 404        & 341        & 252                 \\
Modern   \hphantom & 58        & 10        & 16  \\ \bottomrule
\end{tabular}
\caption{Total number of patients in historic and modern datasets}
\label{dataset_description}
\end{table}

\vspace{-2em}

\subsubsection{Metastases}
Brain metastases occur when cancer cells spread from the primary tumor  to the brain. Brain metastases often cause the leading clinical symptomatology in cancer, therefore their local control is very important. Survival ability of the diseased in case of applying only the supporting therapy amounts to only 40-50 days. The majority of these cases is characterized with multiple lesions, making correct tumor identification and contouring a tedious process. 

\subsubsection{Meningioma}
Most meningiomas are slowly growing, benign neoplasms, deriving from arachnoid cap cells. Meningiomas are usually located on convex, cranial base, cerebral falx and tentorium. A typical meningioma is round-shaped  on the side of the brain and extended on the side of the meninges. As the tumor grows, it may interfere with the normal functions of the brain. The delineation of these tumors is complicated by ``dural tales'' (extended part of the meninges). 

\subsubsection{Schwannoma}
Vestibular schwannomas, or acoustic neuromas, are benign tumors  that arise from the myelin-forming Schwann cells of the vestibulocochlear nerve. As the tumor grows, it presses hearing and vestibular fibres of the auditory nerve  and the  facial nerve, causing hearing loss, tinnitus (ringing in the ear) and loss of balance. If the tumor becomes larger  it can affect  trigeminal nerve and nearby brain structures (such as the brainstem and  the cerebellum,  the fourth ventricle), becoming life-threatening. A typical vestibular schwannoma looks like “comma”, that  arises commonly within the internal auditory meatus, and may extend into the cerebellopontine angle.







\section{Method}
For segmentation, we used standard DeepMedic architecture described in \cite{deepmedic_els}, which is also briefly described in section \ref{sec:related_works}. 
\subsection{Baseline training procedure}
Training procedure, proposed along with DeepMedic architecture \cite{deepmedic_els}, consists of sampling 3D-patches from the images. Central voxels in the first half of the batch are tumorous (foreground) and are healthy (background) in the second half. Therefore, central voxels of the first half are distributed uniformly across all tumorous voxels. This sampling procedure is used to fight class imbalance, since foreground voxels are much rarer than background voxels.

\subsection{Tumor sampling (TS)}

\begin{figure}[h!]
\minipage{0.485\textwidth}
    \includegraphics[width=\linewidth]{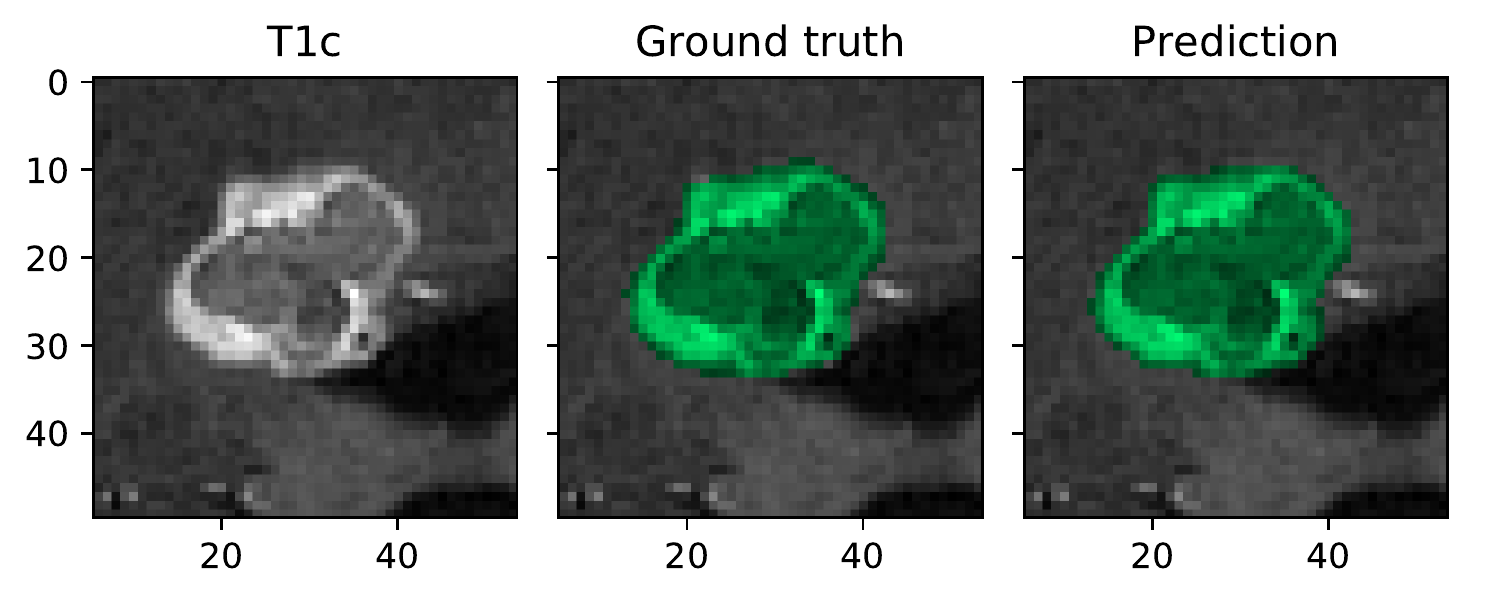}
    \caption{TP prediction of easy-to-detect tumor}\label{fig:TP_big}
\endminipage\hfill
\minipage{0.485\textwidth}
    \includegraphics[width=\linewidth]{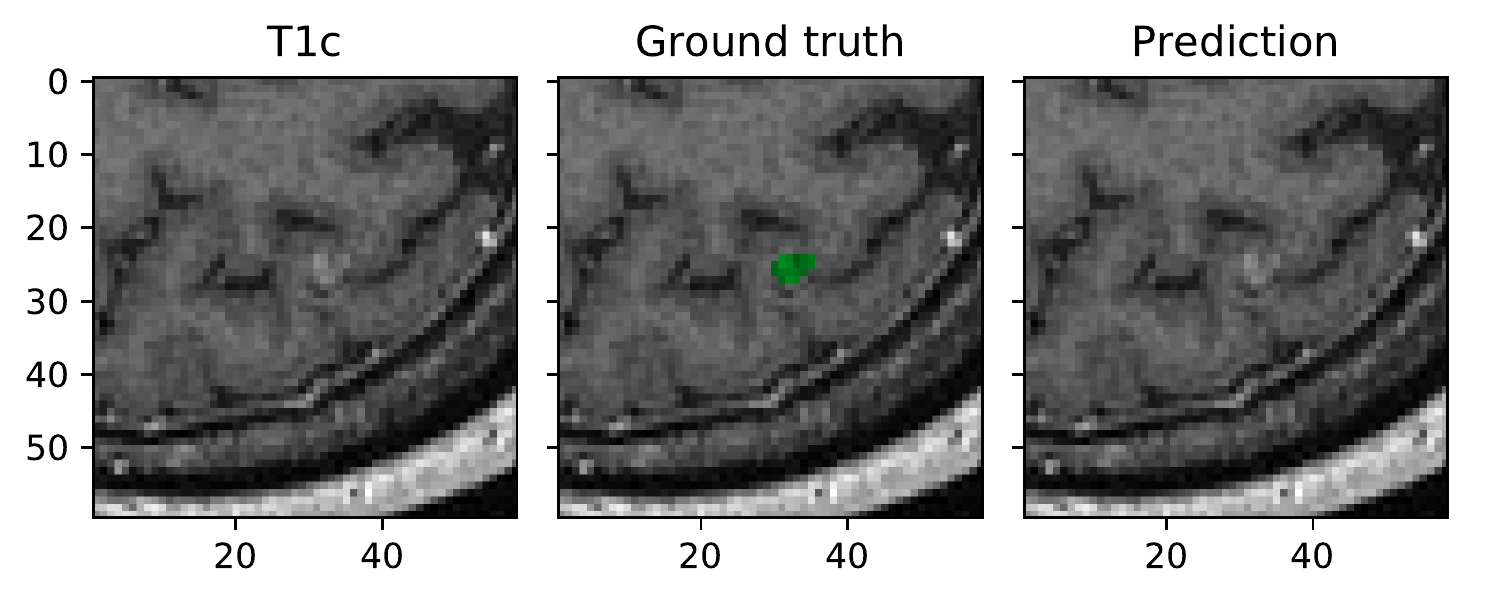}
    \caption{FN prediction of the small tumor}\label{fig:FN_ms}
\endminipage\hfill
\end{figure}

After training our baseline model we discovered that sensitivity was relatively low both in test and train sets. We concluded that the model was not trained well enough. Having observed false negative cases in the training set, we found that many of them were small metastases in a brain which had both big (Figure~\ref{fig:TP_big}) and small (Figure~\ref{fig:FN_ms}) metastases. Original sampling procedure would strongly favour sampling from big metastasis in this case, significantly decreasing number of small metastases observed during training. To fix that we change foreground sampling procedure. Instead of uniformly choosing foreground voxels we first randomly choose a metastasis and only then pick a random voxel inside. This means that now all metastases are equally represented in training set, including the smallest ones. 

\subsection{High intensity sampling (HIS)}

\begin{figure}[h!]
\minipage{0.49\textwidth}
  \includegraphics[width=\linewidth]{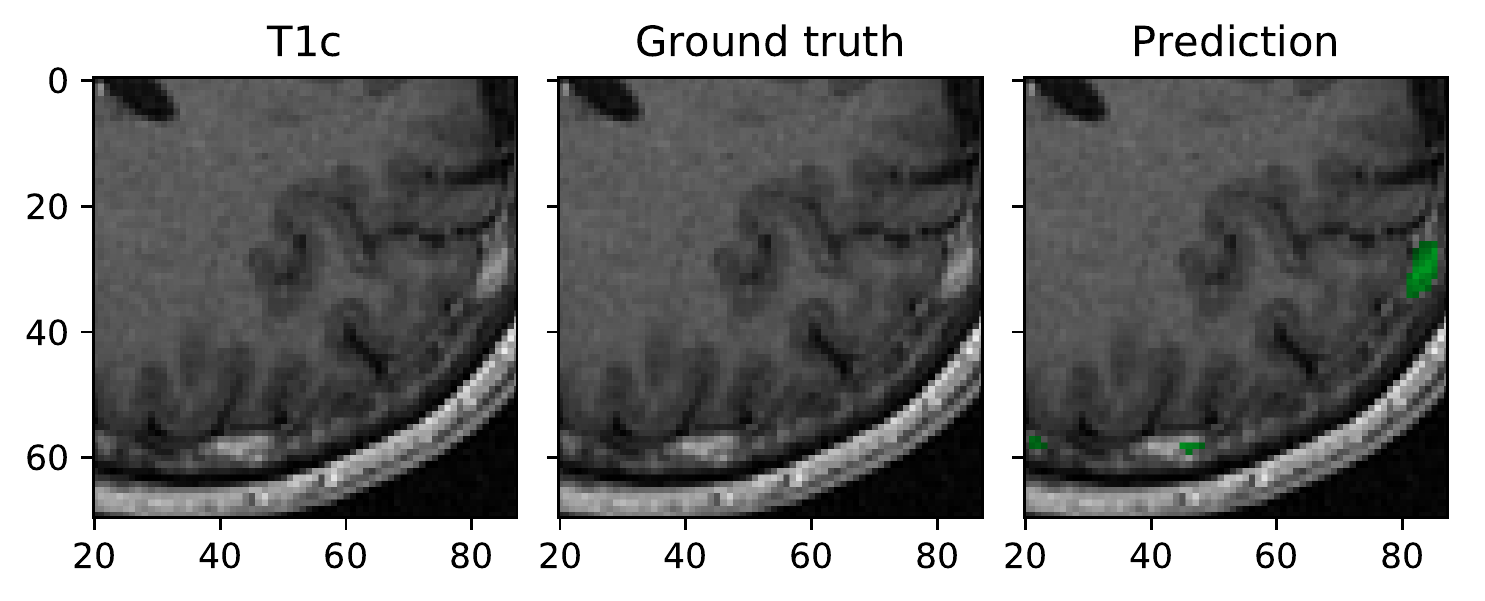}
  \caption{FP prediction of high intensity structures}\label{fig:FP_his}
\endminipage\hfill
\minipage{0.49\textwidth}
  \includegraphics[width=\linewidth]{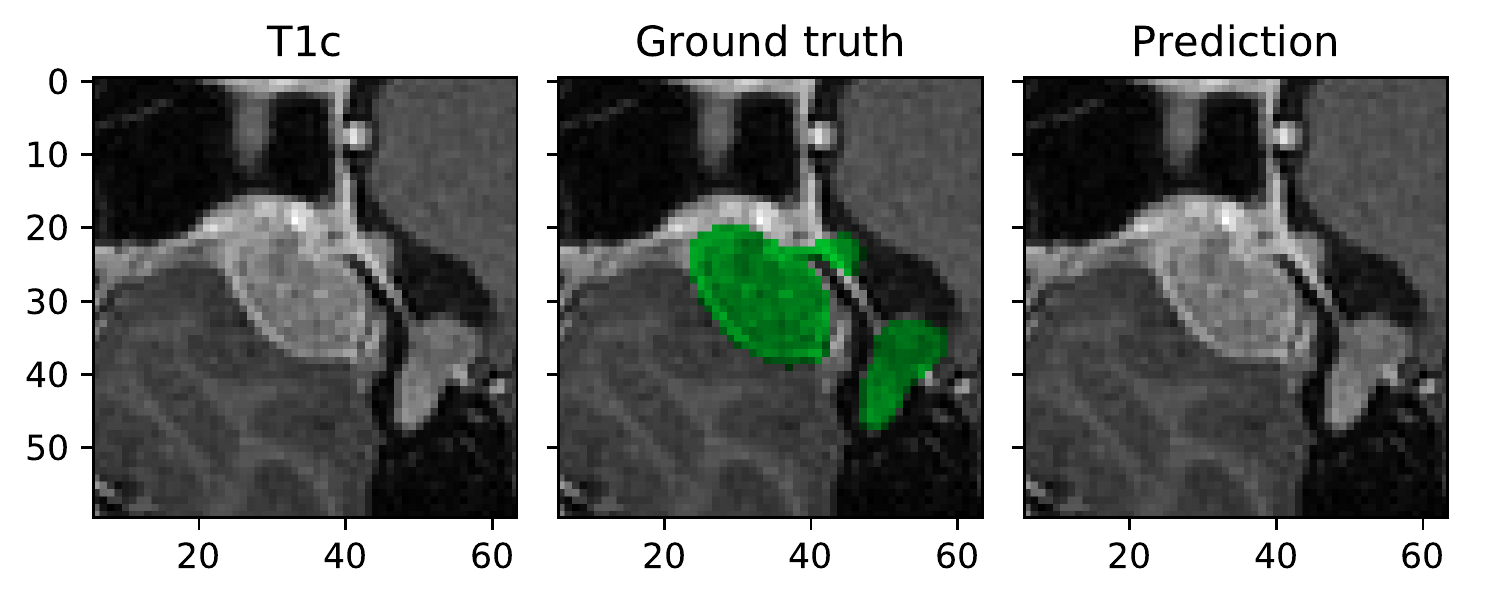}
  \caption{FN predictions of near high intensity structures}\label{fig:FN_his}
\endminipage
\end{figure}

Observing false positive cases we found out that there are many of false predictions in the structures with high intensity level (Figure~\ref{fig:FP_his}) and in some parts of skull. To prevent our model from making such prediction and also to improve predictions near high intensity structures (Figure~\ref{fig:FN_his}), we change sampling procedure almost the same way. Voxels with more than 90-th percentile intensity were chosen as High Intensity class. Then, instead of uniformly choosing a background voxel, we firstly made a decision between High Intensity class (with probability $0.3$) and standard sampling procedure from the background. This approach allows algorithm to learn from hard-to-recognize structures more often.

\section{Experiments}

\subsubsection{Measuring performance}
The tumor delineation process could be considered a tumor detection followed by its segmentation. Detection is a process of spotting the tumors; segmentation is the process of contouring these tumors close to the way the physician did. Detection quality can be measured by using sensitivity and number of false positives computed for tumors. Here we define that tumor was found if there was an intersection between predicted tumor and true tumor. Segmentation quality can be measured by Dice similarity coefficient computed for all patient's voxels. It's a popular metric widely used in segmentation tasks. 


\subsubsection{Training procedure parameters}
During our preliminary experiments we discovered that sampling foreground voxel with probability $0.5$ lead to the large number of false positive examples, so we decreased this probability to $0.25$. During training we are using patch size of $15$ as an output of our model, since this increases sensitivity and dice score in our case. We train our models for 120 epochs since training loss plateaus at this point for any learning rate. Each epoch consists of 200 stochastic gradient descent iterations. We start our training with learning rate of 0.1, and halve it whenever training loss plateaus. 

\subsubsection{Data preprocessing} 
We didn't use standard preprocessing techniques like brain extraction since it can take dozens of minutes. In clinical settings it could annihilate all acceleration of delineation process obtained by deep learning.

\subsubsection{Reproducibility}
We conducted our experiments using Python and PyTorch. Our deep learning algorithms are written in a  highly modular way and fully reproducible thanks to usage of Docker containers. We haven't released it at the moment to preserve anonymity during double-blind reviewing.

\subsection{Results for cross-validation}
First we evaluate our algorithms on each dataset with 5-fold cross validation. Results are presented in Table~\ref{table:historic_results}. We do not apply Tumor Sampling for schwannoma segmentation since most of the corresponding patients have only one tumor. 

\begin{table}[!htb]
    \begin{minipage}{.12\linewidth}
      \centering
        \caption*{}
        \begin{tabular}{|c|}
            \hline
            Setting   \\ \hline
            Baseline  \\ \hline
            TS        \\ \hline
            HIS       \\ \hline
            TS \& HIS \\ \hline
            \end{tabular}
    \end{minipage} 
    \begin{minipage}{.3\linewidth}
      \centering
        \caption*{Metastasis}
\begin{tabular}{|c|c|c|c|}
\hline

  Dice & Sensitivity &    FP \\ \hline

 0.787 &       0.898 &   8.3 \\ \hline
 0.792 &       0.932 &   6.6 \\ \hline
 0.791 &       0.904 &   7.5 \\ \hline
 0.798 &       0.946 &  11.8 \\ \hline

\end{tabular}
    \end{minipage} 
    \begin{minipage}{.27\linewidth}
      \caption*{Meningioma}
      \centering
\begin{tabular}{|c|c|c|c|}
\hline

  Dice & Sensitivity &    FP \\ \hline

 0.735 &       0.929 &   6.9 \\ \hline
 0.737 &       0.933 &   5.6 \\ \hline
 0.731 &       0.925 &   4.9 \\ \hline
 0.742 &       0.928 &   4.6 \\ \hline

\end{tabular}
    \end{minipage}%
    \begin{minipage}{.3\linewidth}
      \centering
        \caption*{Schwannoma}
\begin{tabular}{|c|c|c|c|}
\hline

  Dice & Sensitivity &    FP \\ \hline

 0.881 &       0.975 &   1.6 \\ \hline
   - &         - &   - \\ \hline
 0.874 &       0.979 &   1.7 \\ \hline
   - &         - &   - \\ \hline

\end{tabular}
    \end{minipage} 
    \caption{Results of a 5-fold cross-validation on the historic dataset, TS - Tumor Sampling, HIS - High Intensity Sampling.}
    \label{table:historic_results}
\end{table}





\subsection{Testing on modern data}

\begin{table}[!htb]
    \begin{minipage}{.12\linewidth}
      \centering
        \caption*{}
        \begin{tabular}{|c|}
            \hline
            Setting   \\ \hline
            Baseline  \\ \hline
            TS        \\ \hline
            HIS       \\ \hline
            TS \& HIS \\ \hline
            \end{tabular}
    \end{minipage} 
    \begin{minipage}{.3\linewidth}
      \centering
        \caption*{Metastasis}
\begin{tabular}{|c|c|c|c|}
\hline

  Dice & Sensitivity &    FP \\ \hline

 0.807 &       0.856 &   7.4 \\ \hline
 0.799 &       0.912 &   5.2 \\ \hline
 0.799 &       0.849 &   5.0 \\ \hline
 0.789 &       0.911 &   4.7 \\ \hline

\end{tabular}
    \end{minipage} 
    \begin{minipage}{.27\linewidth}
      \caption*{Meningioma}
      \centering
\begin{tabular}{|c|c|c|c|}
\hline

  Dice & Sensitivity &    FP \\ \hline

 0.615 &       0.927 &   4.3 \\ \hline
 0.604 &       0.927 &   3.4 \\ \hline
 0.697 &       0.927 &   2.5 \\ \hline
 0.685 &       0.936 &   2.4 \\ \hline

\end{tabular}
    \end{minipage}%
    \begin{minipage}{.3\linewidth}
      \centering
        \caption*{Schwannoma}
\begin{tabular}{|c|c|c|c|}
\hline

  Dice & Sensitivity &    FP \\ \hline

 0.793 &       0.962 &   1.8 \\ \hline
   - &         - &   - \\ \hline
   0.8 &       0.962 &   1.4 \\ \hline
   - &         - &   - \\ \hline

\end{tabular}
    \end{minipage} 
    \caption{Experiments on different tumors, TS - Tumor Sampling, HIS - High Intensity Sampling. We used the historic dataset (2006-2011) for training and the modern one (2017) as a hold-out test set to calculate quality metrics}
    \label{table:modern_results}
\end{table}

\begin{figure}[H]

\centering
  \begin{subfigure}[b]{.5\linewidth}
    \centering
    \includegraphics[width=\textwidth]{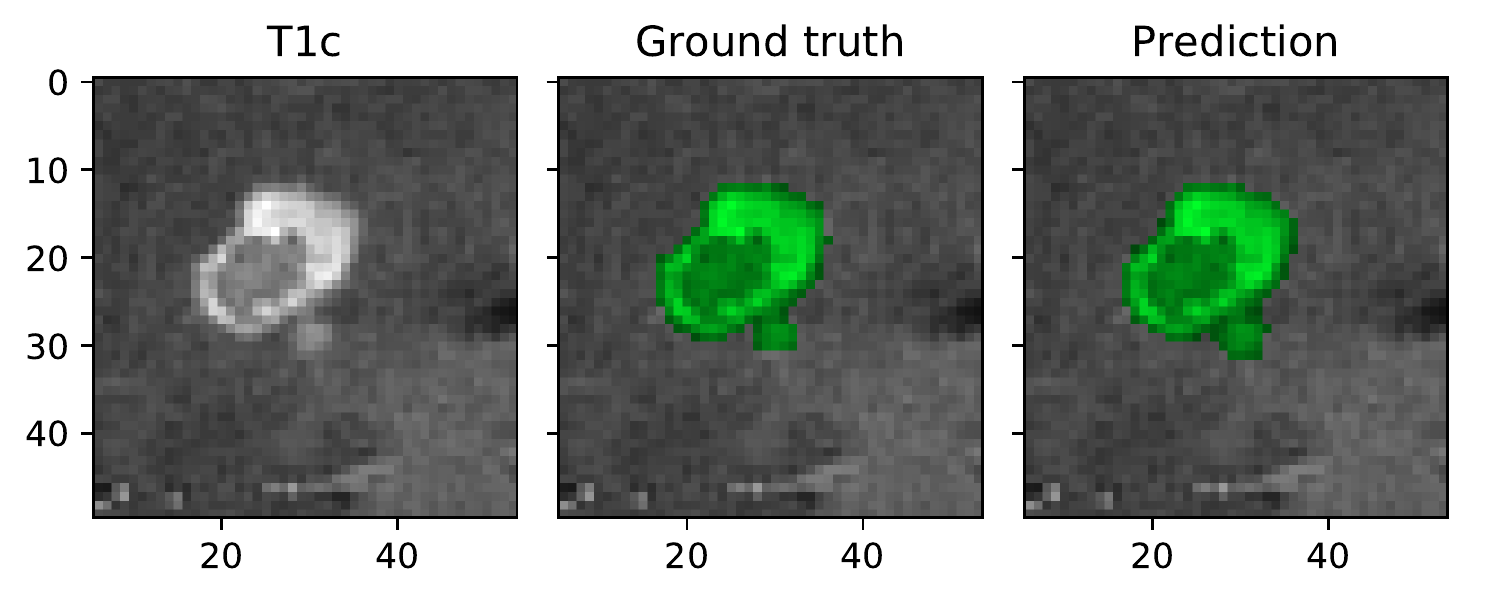}
    \caption{Metastasis segmentation example}\label{fig:examples_met1}
  \end{subfigure}%
  \begin{subfigure}[b]{.5\linewidth}
    \centering
    \includegraphics[width=\textwidth]{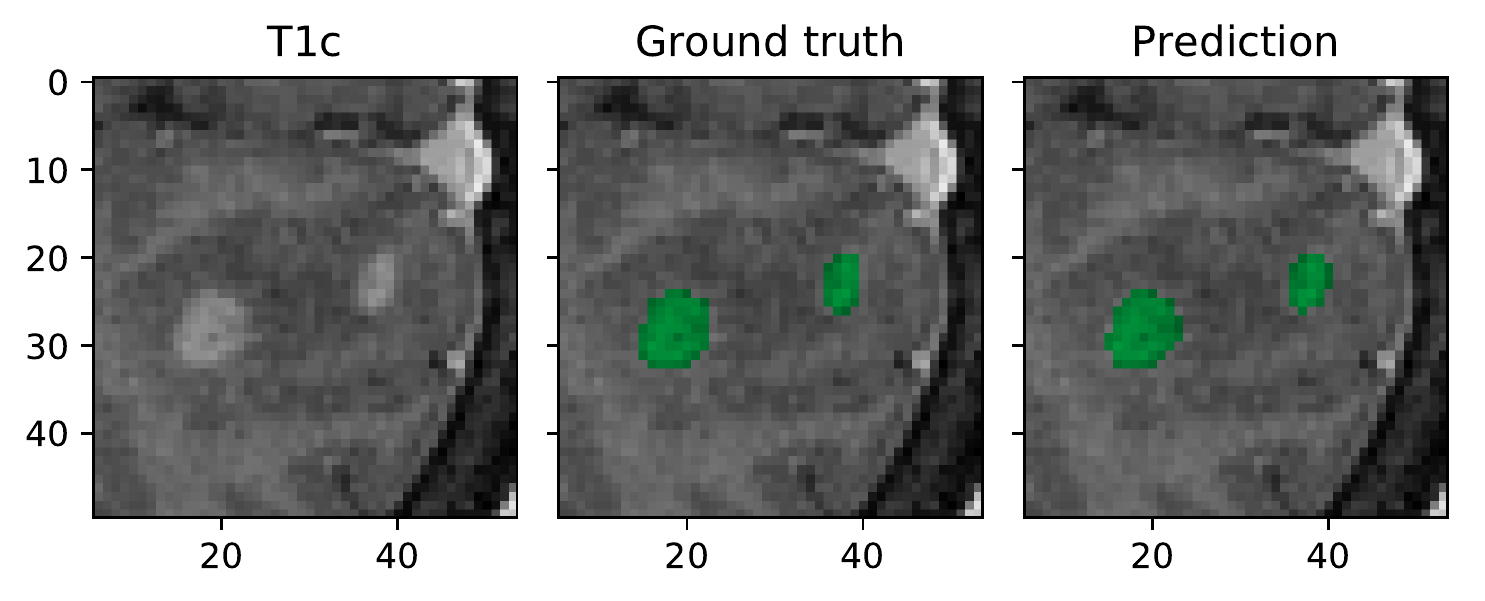}
    \caption{Metastasis segmentation example}\label{fig:examples_met2}
  \end{subfigure}
  \\
  \begin{subfigure}[b]{.5\linewidth}
    \centering
    \includegraphics[width=\textwidth]{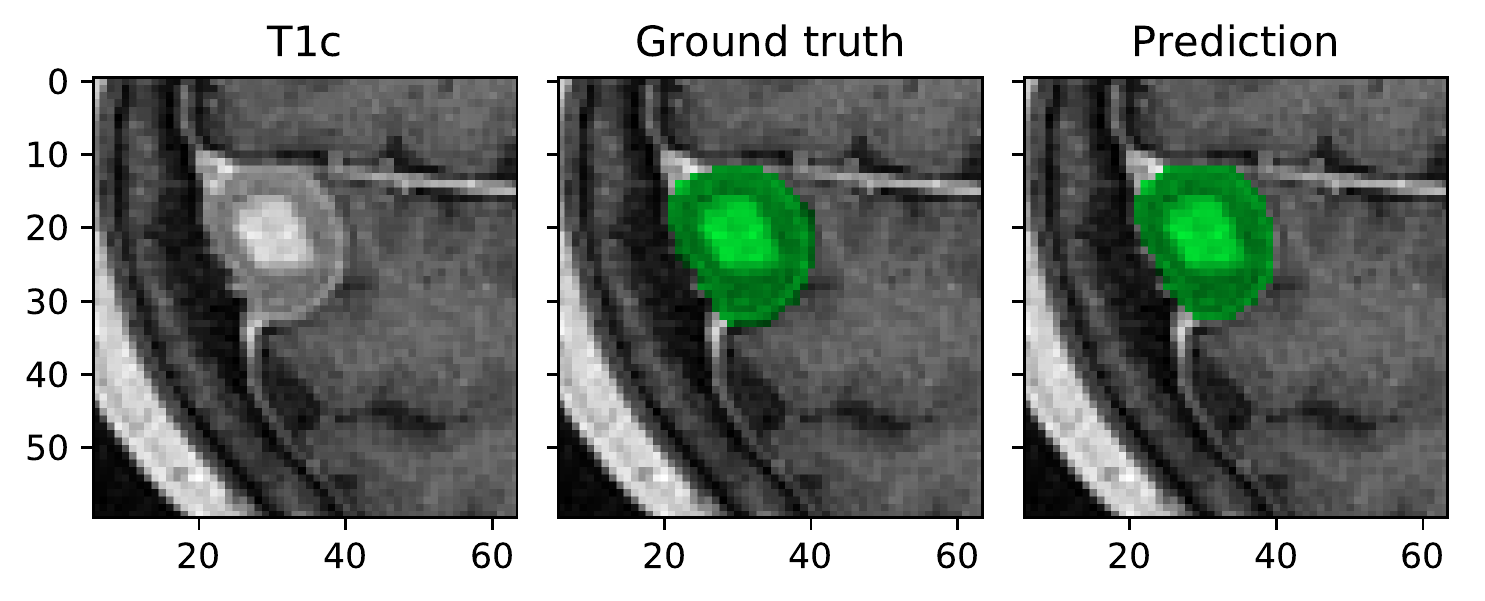}
    \caption{Meningioma segmentation example}\label{fig:examples_menin}
  \end{subfigure}%
  \begin{subfigure}[b]{.5\linewidth}
    \centering
    \includegraphics[width=\textwidth]{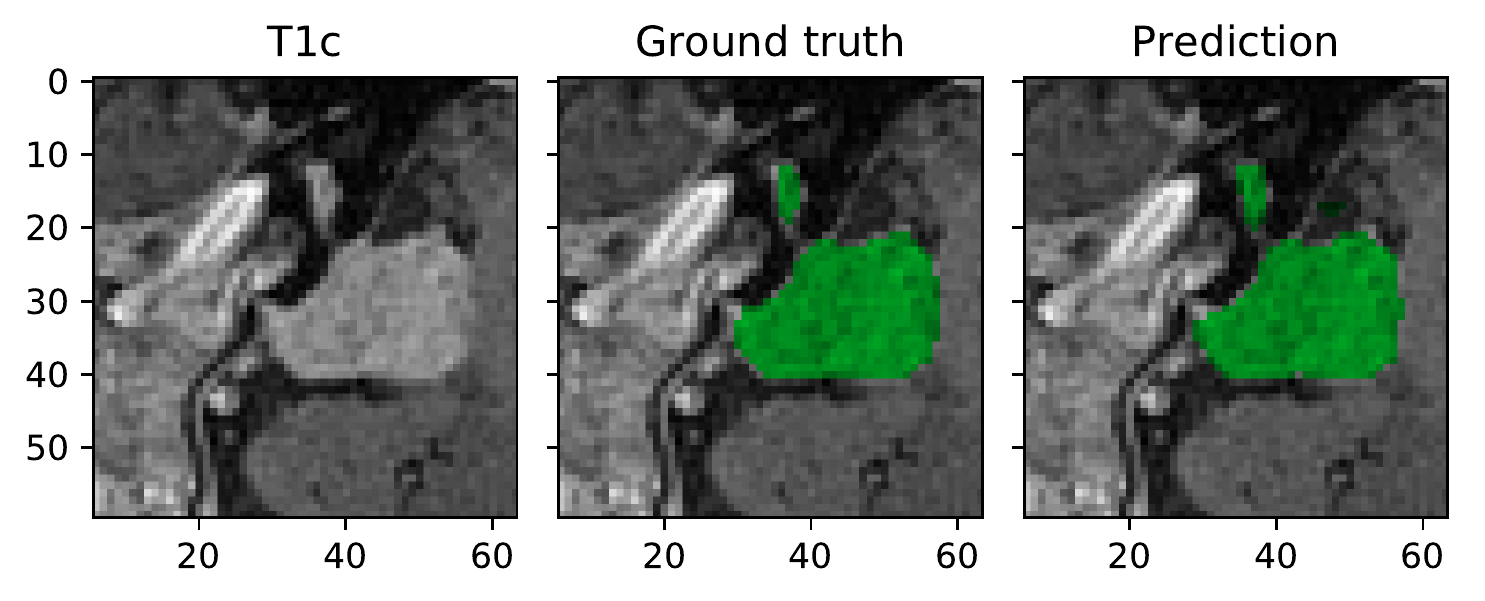}
    \caption{Schwannoma segmentation example}\label{fig:examples_schwa}
  \end{subfigure}   

  \caption{Examples of predicted masks for Tumor Sampling method. Each subfigure contains a T1c image (left), expert annotation (center) and prediction (right) }\label{fig:examples}
\end{figure}

After that we retrain our algorithm on each historic dataset (2006-2011) and test them on modern dataset, which was gathered six years later (2017). Results are presented in Table~\ref{table:modern_results}, see also Figure~\ref{fig:examples} for examples of predicted masks. For meningiona segmentation HIS showed significant increase in dice score. We checked results more precisely: methods differed in how they predicted tumor attached to the skull, with HIS being much more accurate.

These results demonstrates that our algorithm is quite robust to the typical changes of this center, since if we had been able to provide our algorithm six years ago, it would still provide reasonable quality up until this moment. Also, since we could use different algorithms for detection and segmentation, we could always have the best of different models, combining best dice score and sensitivity/FP.

\section{Conclusion}
We developed a new patch sampling strategy  to meet needs of delineating brain lesions for radiosurgery and evaluated the proposed approach by segmenting three of the most common tumors. Also, we emulate long-term usage of our deep learning-based system in clinical settings and demonstrated robust performance of the method.

\section*{Acknowledgements}
The results of sections 1, 2, 4 and 5 are based on the scientific research conducted at IITP RAS and supported by the Russian Science Foundation under grant 17-11-01390.

\bibliographystyle{splncs03}
\bibliography{main}

\end{document}